\documentclass[runningheads]{llncs}

\usepackage{eccv}

\usepackage{multirow}
\usepackage{pifont}
\newcommand{\cmark}{\ding{51}}
\usepackage{makecell}
\usepackage{eccvabbrv}

\usepackage{graphicx}
\usepackage{booktabs}

\usepackage[accsupp]{axessibility}  

\usepackage[breaklinks,colorlinks,citecolor=eccvblue]{hyperref}

\usepackage{orcidlink}
\begin{document}

\title{Pooling-Based Context Modeling for Convolution-Free Deep Image Prior} 


\author{Gihyun Kim
\and
Jong-Seok Lee
}


\institute{Yonsei University\\
Republic of Korea\\
\email{kkh9314@yonsei.ac.kr, jong-seok.lee@yonsei.ac.kr}}

\maketitle

\begin{abstract}
Convolutional Neural Networks (CNNs) achieve strong denoising performance by exploiting spatial context from neighboring pixels. Deep Image Prior (DIP) leverages this property to restore images from a single noisy input without requiring large datasets. However, the over-parameterized architecture of DIP often leads to noise fitting during optimization. In this paper, we propose Pool-DIP, a convolution-free architecture that incorporates pooling-based contrast modeling to capture spatial context efficiently. Pool-DIP improves denoising performance while significantly reducing the number of parameters and computational complexity compared to convolution-based DIP models. Experimental results show that Pool-DIP achieves competitive performance across multiple datasets, including a real-world benchmark. Spectral analysis further reveals that Pool-DIP stabilizes the evolution of high-frequency components during optimization and suppresses erroneous high-frequency signals. The proposed architecture also generalizes well to other image restoration tasks such as super-resolution and inpainting.
\end{abstract}

\section{Introduction}
\label{sec:intro}
Convolution Neural Networks (CNNs) are widely used in computer vision due to their strong performance. A key factor contributing of CNNs is their capability to model \textit{context}, which captures relationships among neighboring pixels. This characteristic makes CNNs particularly effective for image restoration tasks, especially image denoising~\cite{locality04,dataset03}. However, as in other areas of computer vision, CNNs typically require large training datasets to achieve satisfactory denoising performance.  

Deep Image Prior (DIP)~\cite{dip01} reveals that the CNN architecture itself inherently possesses a powerful prior, allowing it to achieve competitive denoising results using only a single noisy image, without the need for extensive datasets. In the absence of large training datasets, the architectural design of DIP becomes particularly important~\cite{dip02,dip03,dip04,search01,spectral01}, as it strongly influences the effectiveness of the resulting prior.



\setlength{\textfloatsep}{12pt}
\begin{figure}[t]
\centering
\includegraphics[width=0.75\linewidth]{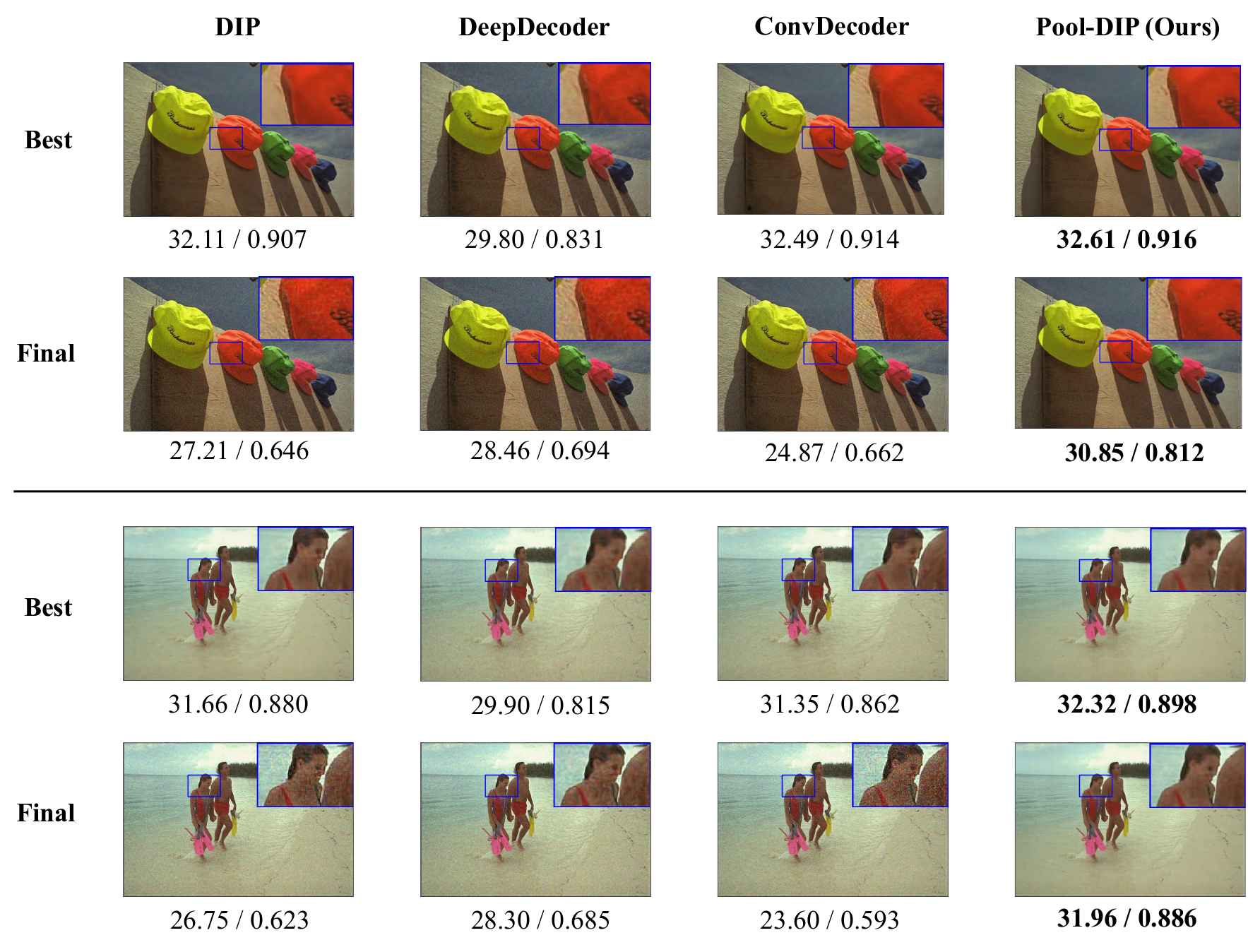}
\caption{Denoising results of DIP, DeepDecoder, ConvDecoder, and our Pool-DIP. The best achievable results during training and the results at the end of training are shown. PSNR (dB) and SSIM are shown below each image. The images are \textit{kodim03} and \textit{kodim12} of the Set9 dataset~\cite{dataset01}.}
\label{figure2}
\end{figure}

An early study~\cite{dip01} used convolution architectures that have a large number of parameters in order to ensure flexibility in the achievable representation space. However, such over-parameterized architectures may lead to fitting noise during optimization~\cite{dip02,overfit01}.

DeepDecoder~\cite{dip02} demonstrates that non-convolutional networks can also serve as DIP architectures when equipped with upsampling layers that enable interactions among neighboring pixels. While this approach reduces noise-fitting by limiting the number of parameters, it also tends to degrade denoising performance since it relies solely on fixed upsampling operations to capture context information, which constrains its expressivity. To address this limitation, ConvDecoder~\cite{dip03} introduces convolution layers into the DeepDecoder architecture. Although this modification improves performance compared to DeepDecoder, it still does not fully match the denoising performance of convolution-based DIP models~\cite{dip01}, and the reduced parameter constraint may weaken its robustness to noise fitting.

Due to the limitation of non-convolutional architectures, several existing studies have focused on how to optimize the performance of convolutional architectures. In particular, it has been attempted to identify optimal network structures via Neural Architecture Search (NAS)~\cite{nas01, nas02, nas03}. However, this approach requires significant computational costs. A recent approach~\cite{search01} simplifies the architectural search process by choosing the network depth and width based on the image characteristics. However, this approach relies on a trained auxiliary model to determine the architecture, introducing additional training overhead and dependency on the auxiliary model.

We observe that under-parameterized non-convolutional architectures with a limited number of learnable parameters can still serve as effective candidates for DIP, as they help mitigate noise fitting without requiring expensive architectural search processes. This raises the following question: \textit{Can a non-convolutional architectural component enable under-parameterization while still capturing sufficient contextual information for effective denoising?} We hypothesize that \textbf{pooling}, by aggregating regional statistics over local neighborhoods, can implicitly capture context information while maintaining an under-parameterized structure. It thereby serves as a lightweight contextual mechanism bridging the gap between upsampling only and convolution-based approaches. Specifically, we propose \textbf{Pool-DIP}, which incorporates pooling operations to enhance contextual modeling within the DIP framework. The core of the proposed model is the \textbf{Multi-Scale Local Contrast Pooling (MS-LCP)} module, which computes the difference between the input features and their pooled representations across multiple scales. Through theoretical analysis, we show that the proposed architecture constrains noise fitting due to its reduced number of spatial parameters. Empirical analysis further reveals that Pool-DIP stabilizes the spectral evolution during optimization, limiting the growth of erroneous high-frequency components compared to the original DIP.



\begin{itemize}
\item We propose Pool-DIP, a pooling-based DIP architecture that achieves competitive denoising performance with under-parameterization. We highlight the importance of contextual modeling in image denoising and show that the proposed MS-LCP module captures spatial relationships across multiple scales.
\item We provide theoretical analysis showing that the proposed architecture constrains noise fitting due to its reduced number of spatial parameters.
\item Through extensive experiments, we demonstrate that Pool-DIP achieves competitive performance while using 32\% fewer parameters than recent DIP architectures across various synthetic datasets and a real-world dataset.
\item We analyze the spectral behavior of DIP during optimization and show that Pool-DIP stabilizes the evolution of high-frequency components while suppressing the growth of erroneous high-frequency signals, providing an explanation for its improved robustness to noise fitting.
\item Additionally, we show that Pool-DIP can be applied to other image restoration tasks beyond denoising, including super-resolution and inpainting.
\end{itemize}

\section{Related Works}
\label{sec:related}
\subsection{Deep Image Prior}
DIP~\cite{dip01} reveals that neural architectures inherently possess a powerful prior, enabling them to produce impressive results for image restoration without training data. Given Gaussian noise as input, the DIP model iteratively reconstructs a denoised image. This phenomenon is closely related to the spectral bias of neural networks~\cite{spectral01,spectral02,spectral03}, which tend to prioritize processing of low-frequency information in the early stages of training.

Improving the performance of DIP has been explored in various ways. Deep RED~\cite{dip05} enhances DIP by integrating an explicit prior. Liu \etal~\cite{dip06} combine total variation regularization with DIP to reduce noise and preserve edges. Deep Random Projector~\cite{dip07} prevents overfitting to noise by making the input learnable, while keeping randomly-initialized network frozen. Wang \etal~\cite{dip08} introduce an early-stopping strategy to capture restored results before overfitting to noise occurs. Sun \etal~\cite{dip09} incorporate a plug-and-play prior scheme, enabling the inclusion of additional regularization steps within the DIP framework.

While applying NAS to DIP has been also attempted~\cite{nas01,nas02,nas03}, this approach is computationally intensive and time-consuming.
A simpler approach for architecture search was proposed in~\cite{search01}. It shows that the upsampling causes the loss of high-frequency content, which could harm the results for fine-grained images. Therefore, starting from a base architecture with only two upsampling layers, layer dimensions are determined through image-specific searches using three trained support vector machines. However, this approach still incurs additional computation in addition to the training process of denoising.

\subsection{Under-Parameterized DIP}
Under-parameterized DIP methods have been proposed to prevent overfitting to noise components in the noisy input image. DeepDecoder~\cite{dip02} uses only 1$\times$1 convolution, reducing the number of parameters. However, its performance is rather limited compared to the original DIP. ConvDecoder~\cite{dip03} has a similar architecture to DeepDecoder but uses 3$\times$3 convolution. With an increased number of parameters, it performs better than DeepDecoder, but is still outperformed by the original DIP, as shown in~\cite{search01}.

Our method also aims to build an under-parameterized model for DIP, where the proposed MS-LCP module enhances contextual modeling by capturing spatial relationships across multiple scales.

\subsection{Unsupervised Single Image Denoising}
Unsupervised single-image denoising methods construct synthetic supervision from a single noisy image. Self2Self~\cite{self2self} uses dropout-based masking to perform self-supervised denoising, while Zero-shot Noise2Noise~\cite{zs_n2n} builds on this idea by proposing a new downsampling method and employing a lightweight convolutional network. Pixel2Pixel~\cite{pixel2pixel} leverages non-local self-similarity within a single noisy image to construct pixel-level pseudo-instances and train a network for zero-shot denoising.

\section{Method}
\label{sec:method}
In this section, we describe the proposed Pool-DIP model. The overall architecture is illustrated in~\cref{figure3}.

\subsection{Capacity-Robustness Trade-off}
\label{sec:tradeoff}
Denoising using DIP is an unsupervised learning task, relying solely on the inductive bias imposed by the structure of CNNs. Consequently, the denoising performance of DIP is significantly influenced by the learning capacity of the network. To motivate our approach, we discuss this influence through the perspective of the trade-off between network capacity and noise robustness. High-capacity architectures provide greater representational flexibility but tend to fit noise more easily, whereas under-parameterized models are more robust to noise fitting but may suffer from limited expressivity.

As shown in~\cref{figure2}, the original DIP~\cite{dip01}, which uses an over-parameterized network having high learning capacity, shows reasonable performance at the best moment but easily overfits to noise eventually. DeepDecoder~\cite{dip02} is under-parameterized with lower capacity and solely depends on upsampling to consider context information, thus it shows higher resistance to noise-fitting but ends up with underfitting to image contents. ConvDecoder~\cite{dip03}, which lies between DIP and DeepDecoder in terms of capacity, improves the representation of image structures compared to DeepDecoder but becomes more susceptible to noise fitting, which can degrade denoising performance.

These observations highlight the trade-off between network capacity and noise robustness in DIP architectures. Upsampling operations alone provide limited ability to capture contextual relationships, while convolution operations with learnable parameters offer high flexibility but may increase the risk of fitting noise. It may be possible to find a network using convolution with a proper balance of the trade-off; however, this requires computationally intensive architectural search~\cite{nas01, nas02, nas03, search01}. We therefore consider pooling as a lightweight mechanism for modeling contextual information while maintaining a relatively low model capacity that helps mitigate noise fitting.

\subsection{Model Architecture}
\subsubsection{Overall Architecture.} Our proposed architecture is a 4-block hourglass network, as illustrated in~\cref{figure3}. Each encoder block gradually compresses spatial resolution while enriching channel representations, and each decoder block symmetrically reconstructs the image through upsampling. Unlike conventional convolution-based DIP architectures, our network models spatial context primarily through pooling operations. The key component is the \textbf{Multi-Scale Local Contrast Pooling (MS-LCP)} module, which is placed in every encoder and decoder block to capture multi-scale spatial relations. The MS-LCP module emphasizes local contrast by comparing input features with their pooled representations at multiple spatial scales. Additionally, positional encoding is incorporated to accelerate the convergence of high-frequency details and inject positional information. Also, we attach a skip path, referred to as the dual-path, to the output of MS-LCP to consider inter-channel relationships simultaneously. 1$\times$1 convolutions are used to model interactions among channels.


\subsubsection{Local Contrast Pooling (LCP).} Let $\mathrm{\mathbf{X}}$ denote the input. Our LCP performs average pooling and subtracts the result from the input $\mathrm{\mathbf{X}}$, i.e., $\mathrm{\mathbf{X}}-\mathrm{Pool(\mathbf{X})}$, and then aggregates the result with $1\times 1$ convolution. 

Pooling aggregates regional statistics over local neighborhoods and therefore provides a lightweight mechanism for modeling spatial relationships among neighboring pixels. In addition, since average pooling smooths the input, it acts as a low-pass filter. Consequently, $\mathrm{\mathbf{X}}-\mathrm{Pool(\mathbf{X})}$ emphasizes high-frequency components important for recovering edges and fine structures. In MS-LCP, this contrast operation is applied at multiple pooling scale. In the decoder, the output of LCP is added with the input $\mathrm{\mathbf{X}}$ by a skip connection. This operation can be interpreted as emphasizing local contrast of features that may become blurred during the upsampling process. As a result, the proposed architecture can better preserve fine structural details during reconstruction.

\subsubsection{Multi-scale LCP (MS-LCP).} By extending LCP, we propose Multi-Scale LCP (MS-LCP), which utilizes multiple pooling window sizes to capture spatial relationships at different scales. We observe that employing only two different pooling sizes already provides noticeable performance improvements; additional experiments are provided in the Supplementary Material. In addition, we introduce learnable coefficient parameters $\alpha_k$ to control the relative contribution of each LCP output corresponding to different pooling scales:
$$\text{MS-LCP}: \text{ChannelMixing}(\Sigma_{k=1}^{n} \alpha_k \times (X-\text{Pool}_{s_k}(X)))$$
where $\text{ChannelMixing}$ denotes 1$\times$1 convolution, $n$ denotes the number of scales in MS-LCP, and $s_k$ is the $k$-th scale of the pooling window.

\setlength{\textfloatsep}{12pt}
\begin{figure}[t!]
\centering
    \includegraphics[width=0.80\textwidth]{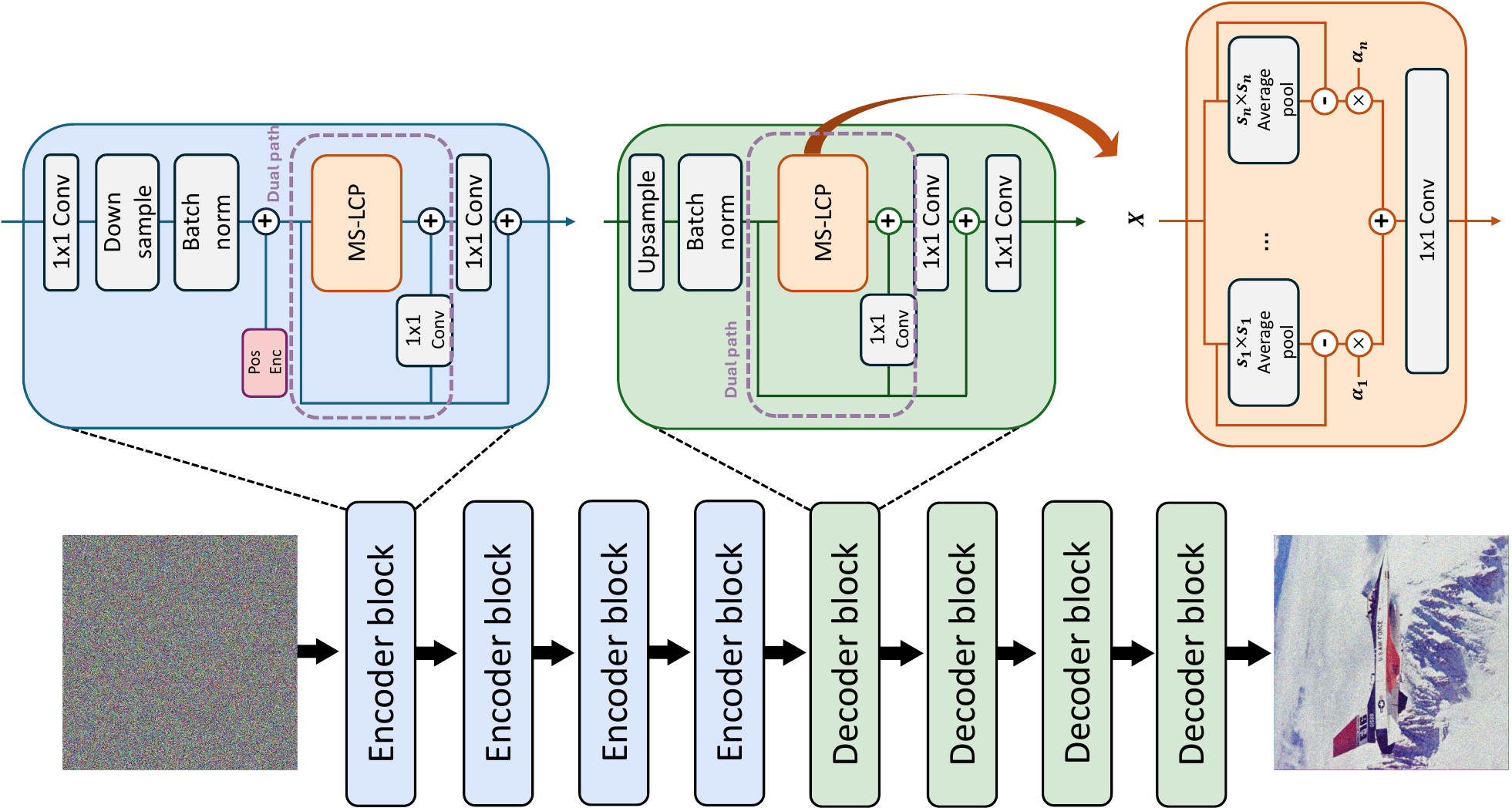}
\caption{Overview of our Pool-DIP architecture. It consists of four encoder blocks and four decoder blocks, each of which contains an MS-LCP module. After each 1$\times$1 convolution, ReLU is applied.}
\label{figure3}
\end{figure}

\subsubsection{Positional Encoding.} Positional encoding is widely used in Transformer architectures to encode positional information in self-attention mechanisms \cite{vit,attention}. Previous studies~\cite{spectral03} have also shown that positional encoding can accelerate the convergence of high-frequency details. In our method, we adopt positional encoding to use this characteristic and help restore high-frequency details. This additional positional information complements the pooling-based contrast modeling in MS-LCP.

\subsubsection{Dual-path Embedding.} In addition to the MS-LCP path for spatial modeling, we introduce a complementary path that captures feature channel interactions. This path is implemented using a $1\times1$ convolution. The outputs of the two paths are combined through summation in each block, as illustrated in~\cref{figure3}:
$$\text{Dual-path}: \text{ChannelMixing}(X)+\text{MS-LCP}(X)$$
This design allows the network to simultaneously capture spatial contrast information through MS-LCP and channel interactions through the dual-path.

\subsection{Mathematical Analysis}
\label{theorem}
In this section, we provide a theoretical analysis of the proposed architecture. Following the analysis framework of~\cite{dip02}, we show that the structure of our network inherently limits its ability to fit noise.

For clarity of analysis, we consider a simplified network consisting of a single decoder block containing one LCP module. Batch normalization is omitted, following the setting in \cite{dip02} and we assume that the output vector has one channel for simplicity. The output of the network is represented as $N(\mathrm{\mathbf{P}})$, i.e.,

\begin{align}
\label{changed_matrix}
N(\mathrm{\mathbf{P}})=\text{ReLU} (\mathrm{\mathbf{QUXC}})\mathrm{\mathbf{w}}.
\end{align}

\noindent Here $\mathrm{\mathbf{P}}=(\mathrm{\mathbf{C}}, \, \mathrm{\mathbf{w}})$ denotes the model parameters, where $\mathrm{\mathbf{C}} \in \mathbb{R}^{k\times k}$ represents a weight matrix of 1 $\times$ 1 convolution in LCP and $\mathrm{\mathbf{w}}$ is a $k$-dimensional vector to produce the output. $\mathrm{\mathbf{Q}} \in \mathbb{R}^{d_1 \times d_1}$ denotes the subtraction between the identity matrix and pooling matrix, $\mathrm{\mathbf{U}} \in \mathbb{R}^{d_1 \times d_0}$ denotes the upsampling matrix, and $\mathrm{\mathbf{X}} \in \mathbb{R}^{d_0 \times k}$ represents the input matrix. Note that the spatial dimension (horizontal and vertical) is merged to one dimension ($d_0$ for input and $d_1$ for output). Under this formulation, the LCP operation can be interpreted as applying a fixed spatial operator that removes locally averaged components from the feature map. Then, the following theorem can be established. 

\begin{quote}
\label{theorem1}
    \textbf{Theorem 1.} Let $\eta \in \mathbb{R}^{d_1}$ denote zero-mean Gaussian noise with covariance matrix $\sigma \mathrm{\mathbf{I}}, \sigma >0$. If $k^2 \log(d_0)/d_1 \le 1/32$, the following inequality holds with probability at least $1-2d_0^{-k^2}$:
    $$\min\limits_{\mathrm{\mathbf{P}}}{||N(\mathrm{\mathbf{P}})-\eta}||^2_2 \ge ||\eta||^2_2 \left( 1-20\frac{k^2 \log(d_0)}{d_1} \right)$$
\end{quote}

The proof is provided in the Supplementary Material. This theorem suggests that a non-convolutional network with LCP has limited ability to approximate noise below a certain error bound. This error bound arises because Pool-DIP introduces no additional learnable spatial parameters beyond the $1\times1$ convolutions, which limits the representational capacity of the network. In particular, the fixed pooling-based spatial operator constrains the spatial degrees of freedom of the network, which contributes to its robustness against noise fitting.

\setlength{\textfloatsep}{12pt}
\begin{table}[t!]
\centering
\caption{Evaluation of the existing and proposed methods on three datasets in terms of PSNR (dB), SSIM, the number of parameters, and computational complexity. The best and second best methods are marked with bold faces and underlines, respectively. Results of other quality metrics can be found in Supplementary Material. $^*$Since different networks are used for different images in the case of Devil \cite{search01}, the average number of parameters is reported.}
\label{table1}
\begin{tabular}{cc|ccccccc} 
\hline \hline
Dataset &  & DIP  & DeepDecoder (DD)  & ConvDecoder (CD) & Devil & Ours \\ 
\hline

\multirow{2}{*}{Set9} & PSNR & \multicolumn{1}{c}{30.21} & \multicolumn{1}{c}{28.73} & \multicolumn{1}{c}{30.03} & \multicolumn{1}{c}{\underline{30.48}} &  \textbf{30.55} \\
                    & SSIM & \multicolumn{1}{c}{\underline{0.890}} & \multicolumn{1}{c}{0.839} & \multicolumn{1}{c}{0.879} & \multicolumn{1}{c}{\underline{0.890}}  & \textbf{0.894} \\
                \hline
\multirow{2}{*}{Set12} & PSNR & \multicolumn{1}{c}{\underline{30.78}} & \multicolumn{1}{c}{29.02} & \multicolumn{1}{c}{30.04} & \multicolumn{1}{c}{30.48}  & \textbf{31.19}    \\
                     & SSIM & \multicolumn{1}{c}{\underline{0.898}} & \multicolumn{1}{c}{0.859} & \multicolumn{1}{c}{0.877} & \multicolumn{1}{c}{0.885} & \textbf{0.900}   \\
                     \hline
\multirow{2}{*}{CBSD68} & PSNR &  \multicolumn{1}{c}{28.77} & \multicolumn{1}{c}{27.63} & \multicolumn{1}{c}{28.58} & \multicolumn{1}{c}{\underline{29.03}} & \textbf{29.10} \\
    & SSIM & \multicolumn{1}{c}{0.819} & \multicolumn{1}{c}{0.609}  & \multicolumn{1}{c}{0.808} & \multicolumn{1}{c}{\underline{0.820}}   & \textbf{0.827}   \\ 
\hline 

\multicolumn{2}{c|}{\# of Parameters} & \multicolumn{1}{c}{2.30M} & \multicolumn{1}{c}{0.10M} & \multicolumn{1}{c}{0.89M}& \multicolumn{1}{c}{$0.34\text{M}^*$} & \multicolumn{1}{c}{0.23M} \\
\hline 
\multicolumn{2}{c|}{\makecell{Computational \\ Complexity (GMac)}} & \multicolumn{1}{c}{7.27} & \multicolumn{1}{c}{6.97} & \multicolumn{1}{c}{59.10}& \multicolumn{1}{c}{34.43} & \multicolumn{1}{c}{\textbf{4.08}} \\

\hline \hline
\end{tabular}
\end{table}

\begin{figure}[t!]
\centering
    \includegraphics[width=\linewidth]{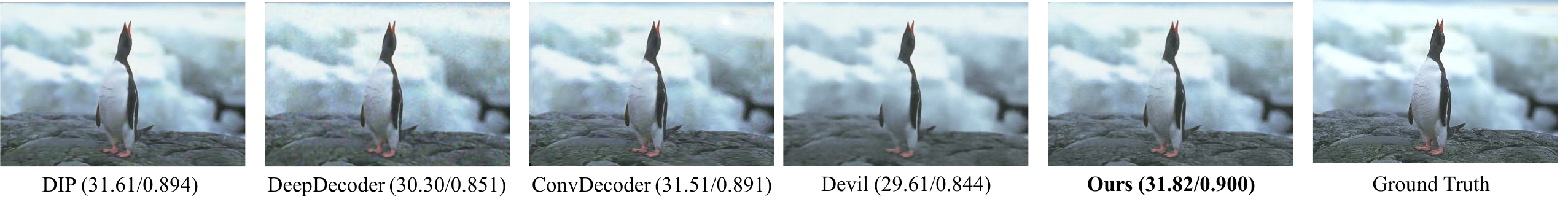}
      \caption{Denoised results for image \textit{0013} of the CBSD68 dataset~\cite{dataset02}. PSNR (dB) and SSIM are shown. Our method is compared to DIP~\cite{dip01}, DeepDecoder~\cite{dip02}, ConvDecoder~\cite{dip03}, and Devil~\cite{search01}.}
\label{figure1}
\end{figure}



\section{Experiments}
\label{section5}
\subsection{Setup}
We follow the experimental recipe of DIP \cite{dip01}. For comparison, we consider the original DIP \cite{dip01}, DeepDecoder~\cite{dip02}, ConvDecoder~\cite{dip03}, and Devil~\cite{search01}, which selects network configurations based on image characteristics. The evaluation is conducted on commonly used datasets, including Set9 \cite{dataset01}, Set12 \cite{dataset03}, and CBSD68 \cite{dataset02}.
By default, we use MS-LCP with two pooling scales (3 and 5) together with sinusoidal positional encoding. We use the Adam optimizer for training our network, where the decay parameter is set to $10^{-7}$. The results of the existing methods are reproduced using our implementation following the settings described in their respective papers. The number of training iterations is set to 5000 for all methods, and the best performance during the training process is recorded for all architectures to compare their maximum achievable performance. More details can be found in Supplementary Material. 



\setlength{\textfloatsep}{12pt}
\begin{table}[t!]
\centering
\caption{Performance comparison under fixed iteration settings.
DIP, DD, CD, and Devil are evaluated at
3000 iterations, while Pool-DIP is evaluated at 1000 iterations due to
its faster convergence. Training time denotes the optimization time per image  measured on a single NVIDIA RTX PRO 6000 Blackwell GPU.}
\label{table1b}
\begin{tabular}{cc|ccccccc} 
\hline \hline
Dataset &  & DIP   & DeepDecoder (DD)   & ConvDecoder (CD)  & Devil  & Ours  \\ 
\hline
\multirow{2}{*}{Set9} & PSNR & \multicolumn{1}{c}{29.59} & \multicolumn{1}{c}{27.67} & \multicolumn{1}{c}{26.80} & \multicolumn{1}{c}{\textbf{29.97}} &  \underline{29.81} \\
                    & SSIM & \multicolumn{1}{c}{\underline{0.863}} & \multicolumn{1}{c}{0.760} & \multicolumn{1}{c}{0.756} & \multicolumn{1}{c}{\underline{0.872}}  & \textbf{0.881} \\
                \hline
\multirow{2}{*}{Set12} & PSNR & \multicolumn{1}{c}{\underline{30.07}} & \multicolumn{1}{c}{26.86} & \multicolumn{1}{c}{25.78} & \multicolumn{1}{c}{29.89}  & \textbf{30.80}    \\
                     & SSIM & \multicolumn{1}{c}{\underline{0.874}} & \multicolumn{1}{c}{0.730} & \multicolumn{1}{c}{0.714} & \multicolumn{1}{c}{0.877} & \textbf{0.890}   \\
                     \hline
\multirow{2}{*}{CBSD68} & PSNR &  \multicolumn{1}{c}{28.20} & \multicolumn{1}{c}{26.64} & \multicolumn{1}{c}{25.91} & \multicolumn{1}{c}{\textbf{28.70}} & \underline{28.43} \\
    & SSIM & \multicolumn{1}{c}{0.793} & \multicolumn{1}{c}{0.701}  & \multicolumn{1}{c}{0.682} & \multicolumn{1}{c}{\underline{0.804}}   & \textbf{0.810}   \\
\hline
\multicolumn{2}{c|}{Training Time (sec)} & \multicolumn{1}{c}{33.36} & \multicolumn{1}{c}{14.52} & \multicolumn{1}{c}{27.24}& \multicolumn{1}{c}{24.86} & \multicolumn{1}{c}{\textbf{13.64}} \\
\hline \hline
\end{tabular}
\end{table}

\setlength{\textfloatsep}{12pt}
\begin{table}[t]
\centering
\begin{minipage}{0.48\textwidth}
\centering

\caption{Denoising performance for Poisson noise on the Set9 dataset.}
\label{table2}
\begin{tabular}{l|lllll} 
\hline \hline
    & \multicolumn{1}{c}{DIP} & \multicolumn{1}{c}{DD} & \multicolumn{1}{c}{CD} & \multicolumn{1}{c}{Devil} &\multicolumn{1}{c}{Ours} \\ 
\hline
PSNR (dB) & \multicolumn{1}{c}{33.31} & \multicolumn{1}{c}{33.88} & \multicolumn{1}{c}{\underline{36.95}}& \multicolumn{1}{c}{32.68}& \multicolumn{1}{c}{\textbf{37.26}} \\
SSIM & \multicolumn{1}{c}{0.945} & \multicolumn{1}{c}{0.951} & \multicolumn{1}{c}{\underline{0.971}}& \multicolumn{1}{c}{0.922}& \multicolumn{1}{c}{\textbf{0.979}}\\
\hline \hline
\end{tabular}

\end{minipage}
\hfill
\begin{minipage}{0.48\textwidth}
\centering
\caption{Effect of MS-LCP, evaluated on the Set9 dataset.} 
\label{table3}
\begin{tabular}{l|ll} 
\hline \hline
    & \multicolumn{1}{c}{PSNR (dB)} & \multicolumn{1}{c}{SSIM}  \\ 
\hline
w/o MS-LCP & \multicolumn{1}{c}{28.25} & \multicolumn{1}{c}{0.829} \\
MS-LCP at encoder & \multicolumn{1}{c}{28.82} & \multicolumn{1}{c}{0.850} \\
MS-LCP at decoder & \multicolumn{1}{c}{30.46} & \multicolumn{1}{c}{0.892} \\

\textbf{MS-LCP at both} & \multicolumn{1}{c}{\textbf{30.55}} & \multicolumn{1}{c}{\textbf{0.894}} \\
\hline \hline
\end{tabular}

\end{minipage}

\end{table}

\subsection{Performance Comparison}

\cref{table1} presents quantitative comparisons on the three datasets together with the number of parameters and the model complexity, and~\cref{figure1} shows visual examples of the restored images. Our architecture consistently achieves competitive denoising performance across all datasets while maintaining fewer parameters and low computational complexity. 

Compared to the best existing method, i.e., Devil, our method achieves slightly better performance while using 32\% fewer parameters and lower computational cost. Compared to DIP, our method achieves improved denoising performance while using 90\% fewer parameters. Although DeepDecoder has fewer parameters, it exhibits higher computational complexity due to the use of large channel sizes in every block. A similar trend is observed for Devil, whose channel sizes are adaptively determined based on the frequency characteristics of the input image; as a result, the selected architecture may still involve relatively large channel widths, which increases the computational cost. ConvDecoder shows the highest computational complexity because of its convolution operations combined with large channel sizes similar to those used in DeepDecoder.

\begin{figure}[t!]
\centering
  \begin{subfigure}{0.48\textwidth}
  \centering
  \includegraphics[width=\textwidth]{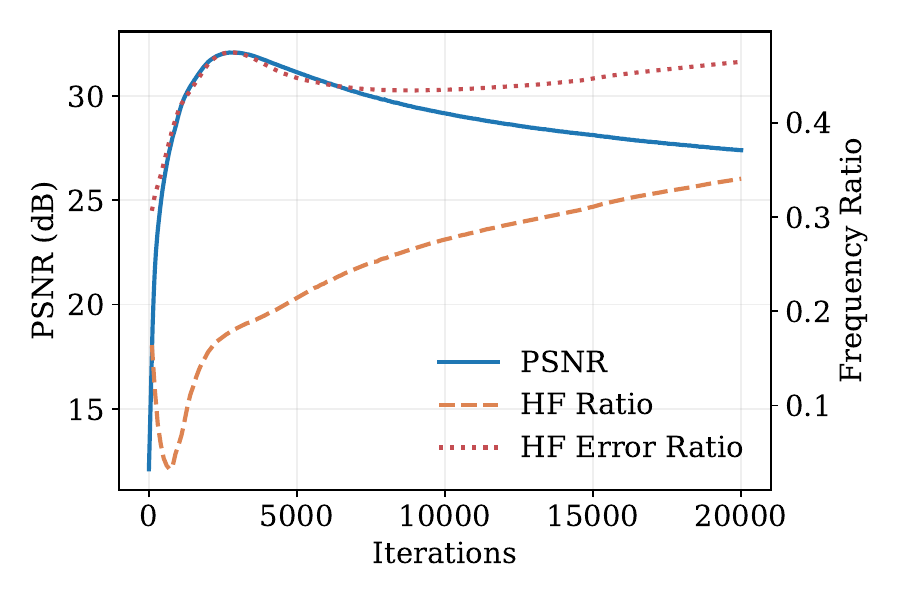}
  \caption{DIP}
  \end{subfigure}
  \begin{subfigure}{0.48\textwidth}
    \includegraphics[width=\textwidth]{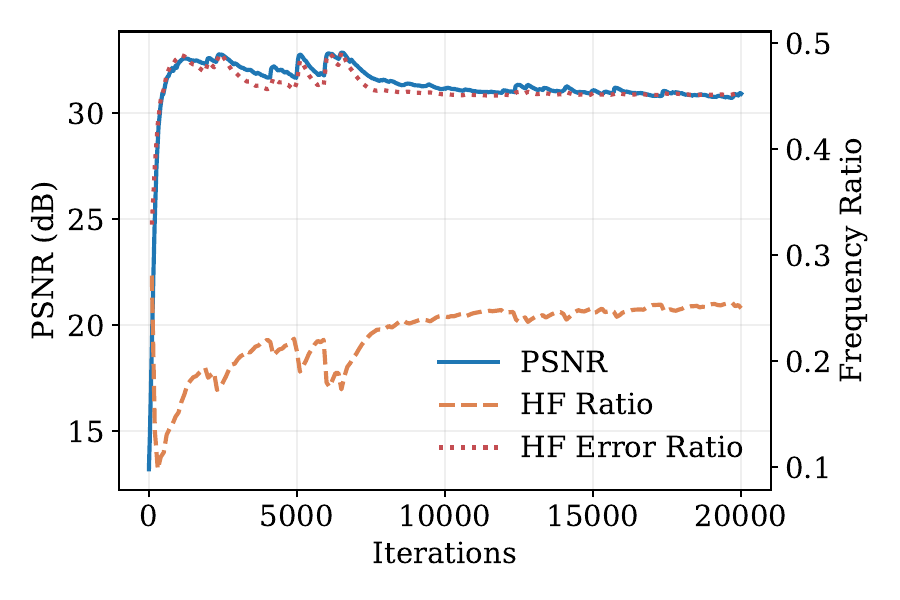}
  \caption{Pool-DIP (Ours)}
  \end{subfigure}
\caption{Optimization dynamics of DIP and Pool-DIP for \textit{kodim03} of the Set9 dataset~\cite{dataset01}. PSNR, high-frequency (HF) ratio of the reconstructed image, and HF error ratio of the reconstruction error during optimization.}
\label{figure5b}
\end{figure}

\cref{table1b} further reports the results under fixed iteration settings. DIP, DeepDecoder, ConvDecoder, and Devil are evaluated at 3000 iterations, and Pool-DIP is evaluated at only 1000 iterations due to its faster convergence (as shown in~\cref{figure5b}). Despite using substantially fewer optimization steps, Pool-DIP achieves competitive or superior performance across the datasets. In addition, Pool-DIP requires the shortest training time per image, demonstrating that the proposed architecture not only improves denoising performance but also provides efficient optimization behavior.

These results demonstrate that the proposed non-convolutional architecture with MS-LCP achieves a favorable balance between denoising performance and computational efficiency.

Additionally we evaluate the models under Poisson noise in~\cref{table2}. Pool-DIP achieves the best performance among the compared methods under Poisson noise.




\setlength{\textfloatsep}{12pt}

\subsection{Optimization Dynamics Analysis}
We examine the optimization behavior by jointly analyzing PSNR, the high-frequency (HF) ratio of the reconstructed image, and the high-frequency (HF) error ratio of the reconstruction error during training. The HF ratio is computed from the magnitude spectrum of the reconstructed image obtained via the 2D discrete Fourier transform, where the ratio of summed magnitudes in the higher half frequency range is measured. As shown in~\cref{figure5b}, the original DIP continuously increases the HF ratio of the reconstructed image as optimization progresses. However, the HF ratio of the reconstruction error also increases in later iterations, suggesting that many of these high-frequency components correspond to reconstruction errors rather than meaningful image structures. This behavior reflects the well-known noise-fitting phenomenon of DIP during prolonged optimization.

In contrast, the proposed Pool-DIP exhibits a more stable optimization behavior. While the reconstructed image gradually accumulates high-frequency components, the growth of the HF ratio is slower and the HF ratio of the reconstruction error remains relatively stable throughout optimization. This observation suggests that MS-LCP limits the growth of error for high-frequency components and results in a more stable spectral evolution during optimization, thereby reducing the tendency of DIP to fit noise during later iterations. This difference in spectral behavior highlights the role of MS-LCP in controlling the generation of high-frequency components during optimization.

\subsection{Analysis of LCP}
\begin{figure}[t]
\centering
\begin{minipage}[t]{0.56\textwidth}
\centering
  \raisebox{0.03cm}{\includegraphics[width=\textwidth]{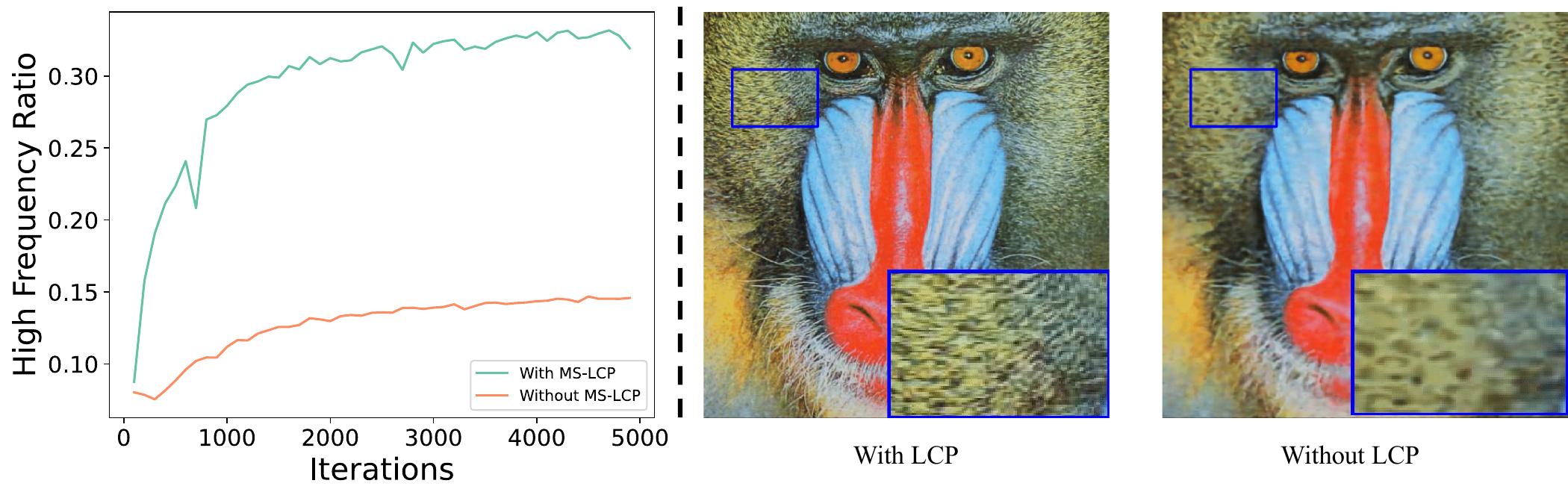}}
\captionof{figure}{(Left) Ratio of the high-frequency components in the restored image during training with and without MS-LCP. (Right) Visualization of the effect of MS-LCP.}
\label{figure5}
\end{minipage}
\hfill
\begin{minipage}[t]{0.40\textwidth}
\centering
\includegraphics[width=\linewidth]{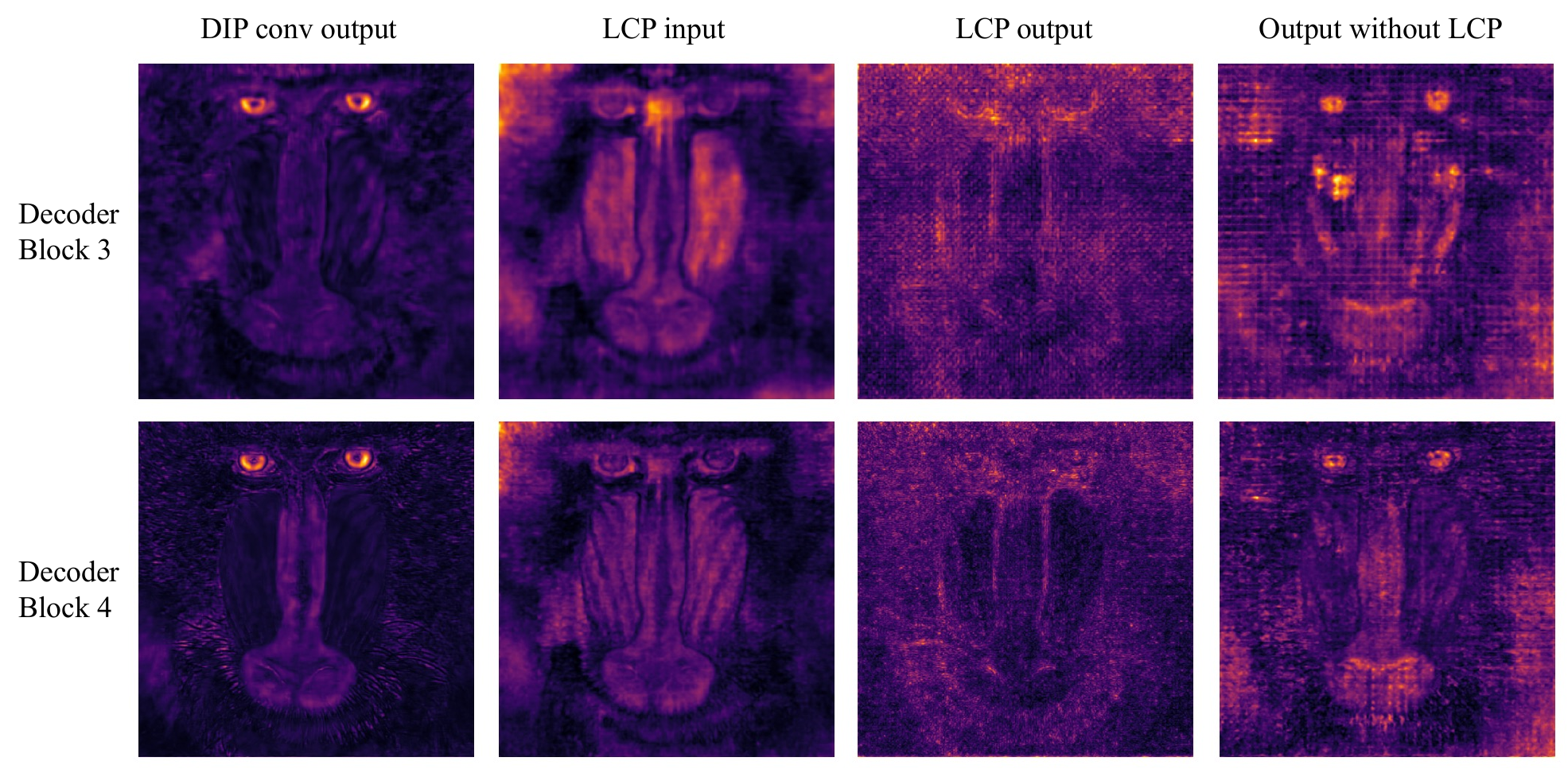}
\captionof{figure}{Visualization of activation maps of the convolution layer in DIP, the input to LCP ($X$), the output of LCP ($X-\text{Pool}(X)$), and the output of MS-LCP without LCP.}
\label{figure4}
\end{minipage}
\end{figure}

To gain further insight into the role of MS-LCP, we analyze the spectral characteristics of reconstructed images during training. \cref{figure5} (left) shows the evolution of HF ratio during optimization. Without MS-LCP, the reconstructed images tend to contain smaller high-frequency components, which is associated with degraded denoising performance, as reflected in~\cref{figure5} (right).

To qualitatively illustrate the effect of MS-LCP, we visualize intermediate activation maps from the last two decoder blocks, as shown in~\cref{figure4}. Compared with convolutional features in DIP, LCP responses emphasize sharper local structures. When LCP is removed, the activations become spatially less coherent and lose fine edge structures, indicating the role of LCP in enhancing local contrast.



We also provide a quantitative comparison in~\cref{table3}. A non-convolutional network without MS-LCP shows a noticeable drop in performance. While MS-LCP improves performance when applied in either the encoder or decoder, the improvement is more pronounced when it is placed in the decoder. This observation suggests that MS-LCP contributes to extracting high-frequency content information (e.g., edges) during the upsampling process in the decoder.

These results indicate that pooling-based contrast modeling helps enhance local contrast and stabilizes the reconstruction process. Without MS-LCP, restoring high-frequency details becomes more difficult due to the absence of both contextual modeling and a mechanism for emphasizing local contrast.

\setlength{\textfloatsep}{12pt}

\subsection{Real-world Dataset}
\label{real-world}
We additionally evaluate our method on the real-world dataset, PolyU~\cite{polyu}, which has 100 noisy-clean image pairs. In this setting, we also provide comparisons with representative zero-shot single-image denoising methods: Self2Self~\cite{self2self}, Zero-shot Noise2Noise~\cite{zs_n2n}, and Pixel2Pixel~\cite{pixel2pixel}. All results are reproduced using the publicly available source codes. Devil~\cite{search01} determines the channel width for each image using a frequency-based classifier. Since this component is not available in the public implementation, we instead use its 64-channel variant, which has a similar number of parameters to our model. As shown in~\cref{table5}, Pool-DIP achieves the best performance among the compared methods, while several existing approaches perform comparably to or worse than DIP in this setting. Among the zero-shot denoising methods, Self2Self achieves the highest PSNR but requires multiple forward passes, while Pixel2Pixel requires additional preprocessing. ZS-N2N provides faster inference but yields lower denoising performance. This result suggests that the proposed architecture generalizes well to real-world noise beyond synthetic noise settings.

\setlength{\textfloatsep}{12pt}
\begin{table}[t!]
\centering
\caption{Denoising performance of our method on the real-world PolyU dataset compared to DIP~\cite{dip01}, Devil~\cite{search01}, and zero-shot single-image denoising methods, including Self2Self (S2S)~\cite{self2self}, Zero-shot Noise2Noise (ZS-N2N)~\cite{zs_n2n}, and Pixel2Pixel (P2P)~\cite{pixel2pixel}.} 
\label{table5}
\begin{tabular}{l|llllll} 
\hline \hline
    & \multicolumn{1}{c}{DIP} & \multicolumn{1}{c}{Devil} & \multicolumn{1}{c}{S2S} & \multicolumn{1}{c}{ZS-N2N} & \multicolumn{1}{c}{P2P} & \multicolumn{1}{c}{Ours} \\ 
\hline
PSNR (dB) & \multicolumn{1}{c}{\underline{38.34}} & \multicolumn{1}{c}{37.45} & \multicolumn{1}{c}{37.70}& \multicolumn{1}{c}{35.84}& \multicolumn{1}{c}{37.34}& \multicolumn{1}{c}{\textbf{38.42}} \\
SSIM & \multicolumn{1}{c}{\underline{0.973}} & \multicolumn{1}{c}{0.961} & \multicolumn{1}{c}{0.956}& \multicolumn{1}{c}{0.937}& \multicolumn{1}{c}{0.960}& \multicolumn{1}{c}{\textbf{0.974}}\\
\hline \hline
\end{tabular}
\end{table}

\subsection{Ablation Study}
We conduct additional experiments to analyze the impact of key design choices in our Pool-DIP architecture. All experiments are conducted on the Set9 dataset.

\subsubsection{Positional Encoding, Dual-path, Coefficient \texorpdfstring{$\alpha$}{alpha}.} 
\cref{table8} highlights the impact of positional encoding, Dual-path, and learnable coefficient $\alpha$. When all components are removed, the overall performance is decreased noticeably. Dual-path (i.e., the residual connection) improves performance, particularly in SSIM, as it enables the network to capture image characteristics through learnable channel interactions via 1$\times$1 convolution. Employing $\alpha$ provides the most noticeable improvement, as it aggregates the outputs of all LCP scales while weighting them according to their relative importance. Concatenation could serve as an alternative approach, but it increases the number of parameters. 

Positional encoding further improves performance, as it accelerates the convergence of high-frequency details (see more details in the next subsection) and injects positional information into the network. As shown in \cref{figure7} (left), positional encoding aids in recovering high-frequency details.

\cref{figure6} shows the average values of $\alpha$ for two representative images from the CBSD68 dataset: \textit{0047}, which predominantly contains low-frequency content, and \textit{0052}, which exhibits richer high-frequency details. We observe that the pooling scale with larger kernel size (5) receives higher weight ($\alpha_1 < \alpha_2$) in \textit{0047}, while the smaller pooling size (3) is more emphasized in \textit{0052}. This observation suggests that the network tends to emphasize different spatial scales depending on the frequency characteristics of the input.


\setlength{\textfloatsep}{12pt}
\begin{table}[t]
\centering
\begin{minipage}{0.48\textwidth}
\centering
\caption{Ablation on positional encoding (PE), Dual-path, and coefficient $\alpha$ on Set9.}
\label{table8}
\begin{tabular}{ccc|cc}
\hline \hline
PE & Dual-path & $\alpha$ & \multicolumn{1}{c}{PSNR (dB)} & SSIM \\
\hline
- & - & - &  29.48 & 0.874 \\
- & - & \cmark & 30.28 & 0.887 \\
- & \cmark & - & 29.83 & 0.880 \\ 
- & \cmark & \cmark & 30.23 & 0.886 \\
\cmark & - & - & 30.02 & 0.885 \\
\cmark & - & \cmark & 30.29 & 0.888 \\
\cmark & \cmark & - & 30.15 & 0.890 \\
\cmark & \cmark & \cmark &
\textbf{30.55} & \textbf{0.894} \\
\hline \hline
\end{tabular}
\vspace*{0.4cm}
\centering
\caption{Performance depending on the location of the positional encoding.}
\label{table9}
\begin{tabular}{l|ll} 
\hline \hline
    & \multicolumn{1}{c}{PSNR (dB)} & \multicolumn{1}{c}{SSIM}  \\ 
\hline
In both & \multicolumn{1}{c}{30.10} & \multicolumn{1}{c}{0.884} \\

In decoder & \multicolumn{1}{c}{29.93} & \multicolumn{1}{c}{0.880} \\
\textbf{In encoder} & \multicolumn{1}{c}{\textbf{30.55}} & \multicolumn{1}{c}{\textbf{0.894}}  \\
\hline \hline
\end{tabular}
\end{minipage}
\hfill
\begin{minipage}{0.48\textwidth}
  \includegraphics[width=\textwidth]{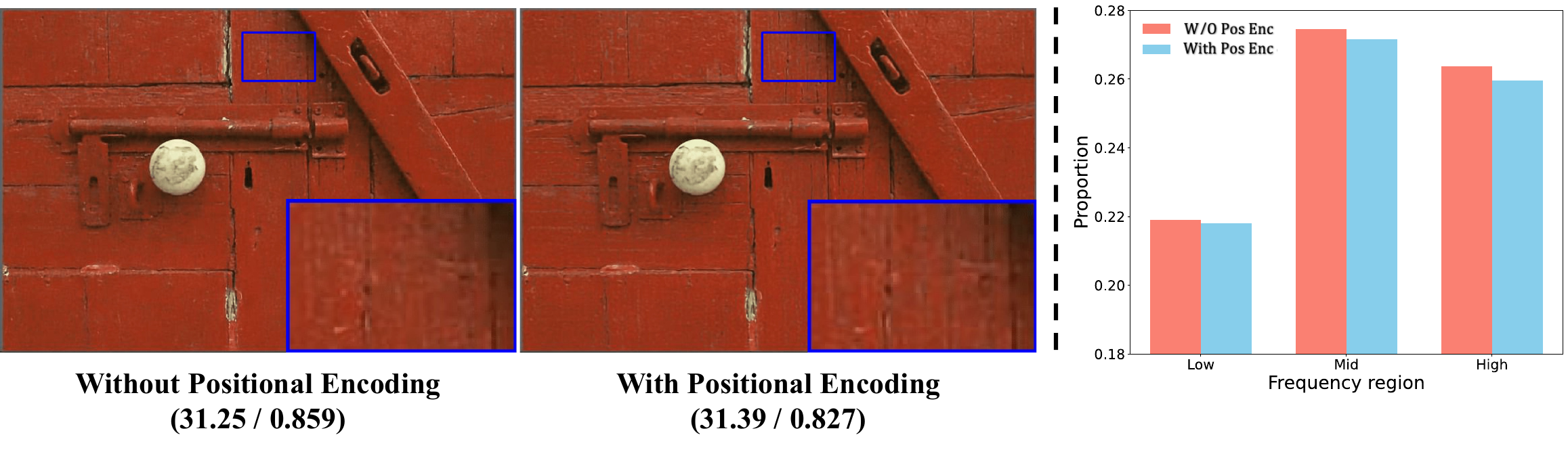}
\captionof{figure}{(Left) Visualization of the effect of positional encoding. PSNR (dB) and SSIM are also shown. (Right) Impact of positional encoding across frequency
regions for \textit{kodim12} in the Set9 dataset.}
\label{figure7}

\centering

 \includegraphics[width=\textwidth]{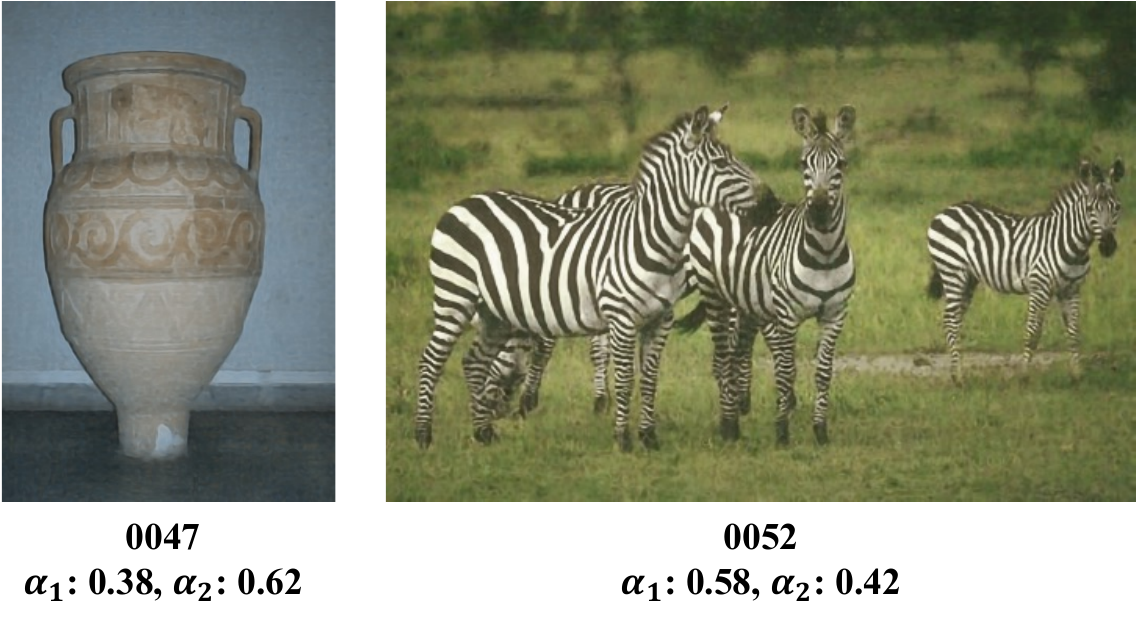}
 \captionof{figure}{Learned values of $\alpha$ for images containing different frequency characteristics.}
\label{figure6}
\end{minipage}
\end{table}




\subsubsection{Frequency Analysis of Positional Encoding.} We analyze the impact of positional encoding across different frequency bands in the reconstruction error. Specifically, we compute the reconstruction error—the difference between the ground-truth and denoised images—and decompose it into low, mid, and high-frequency components based on the Fourier magnitude spectrum. As shown in~\cref{figure7} (right), including positional encoding noticeably reduces the reconstruction error in the high-frequency band. This suggests that positional encoding helps the network recover fine image details, which are often lost during denoising.

 \subsubsection{Location of Positional Encoding.} We also evaluate the performance when positional encoding is applied exclusively to the decoder or simultaneously to both encoder and decoder. As shown in~\cref{table9}, positional encoding is not beneficial when it is applied only to the decoder. This may occur because introducing additional high-frequency components only during the restoration stage encourages the network to generate high-frequency signals that are not sufficiently constrained by earlier feature representations, making the model more susceptible to noise fitting.

\subsection{Other Restoration Tasks}
To evaluate the applicability of Pool-DIP beyond denoising, we conduct experiments on two additional image restoration tasks: super-resolution and image inpainting. We incorporate an additional pooling scale (7) to capture contextual information over larger spatial regions, which is particularly useful for these restoration tasks. For Devil, we use the 64-channel variant as in~\cref{real-world} for a fair comparison.

\subsubsection{Super-resolution.}
\cref{table10} shows the results of DIP, Devil, and our Pool-DIP on the Set14 dataset~\cite{set14}. Our Pool-DIP achieves competitive performance, showing the highest or second-highest performance across both super-resolution scales. Devil also shows strong performance; however, its SSIM for $2\times$ upscaling is lower than the other methods.

\subsubsection{Image Inpainting.}
\begin{figure}[t]
\centering
\begin{minipage}[t]{0.50\textwidth}
\centering
    \includegraphics[width=\textwidth]{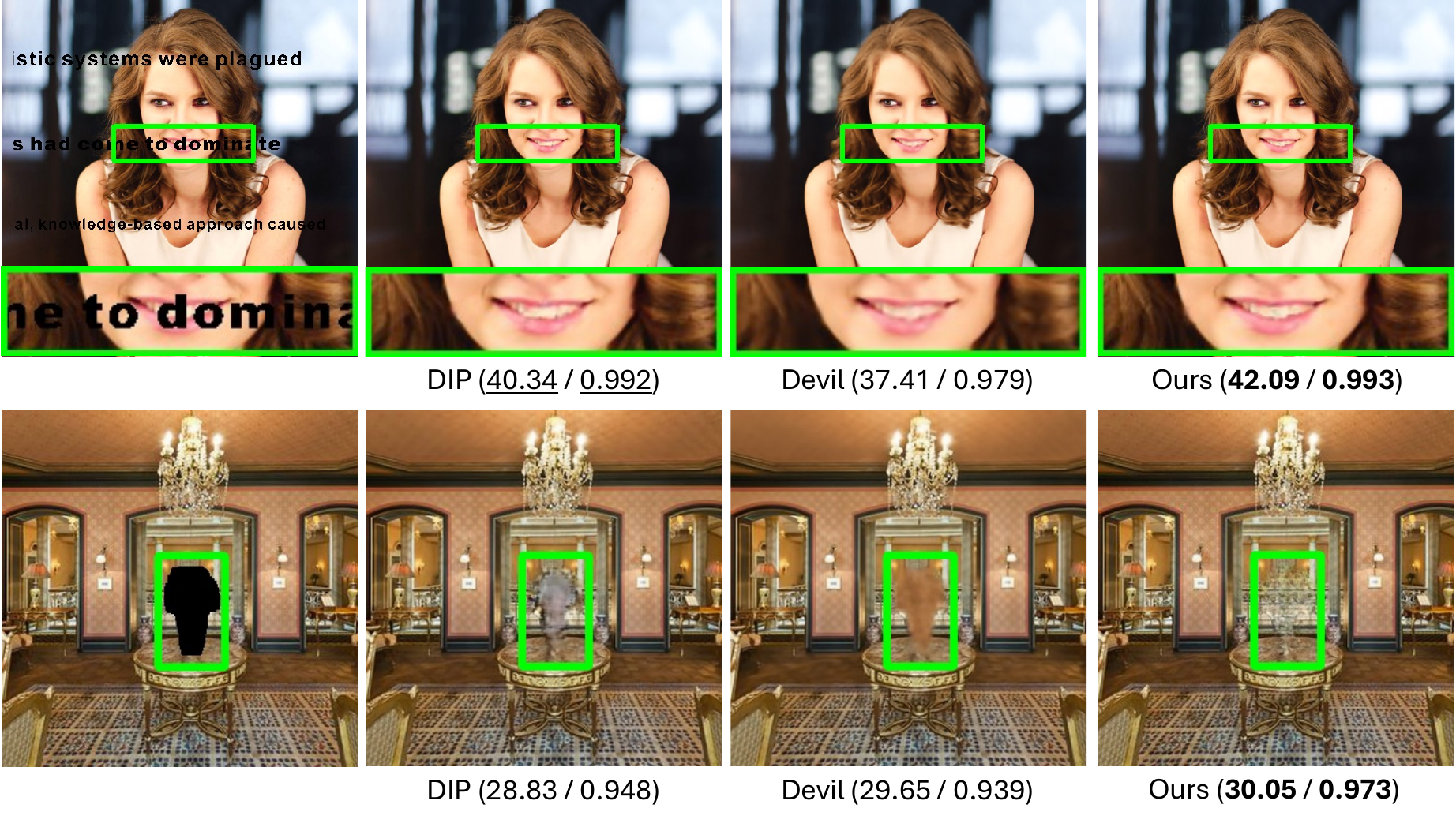}
\caption{Results of image inpainting.}
\label{figure8}
\end{minipage}
\hfill
\begin{minipage}[t]{0.47\textwidth}
\centering
\vspace*{-3.5cm}
\captionof{table}{Performance of super-resolution on the Set14 dataset.}
\label{table10}
\begin{tabular}{cc|ccccc} 
\hline \hline
 Scale &  & DIP & Devil &  Ours \\ 
\hline

\multirow{2}{*}{$2\times$} & PSNR (dB) & \multicolumn{1}{c}{29.04} & \multicolumn{1}{c}{\textbf{29.08}} & \multicolumn{1}{c}{\underline{29.05}}  \\
                    & SSIM & \multicolumn{1}{c}{\underline{0.934}} & \multicolumn{1}{c}{0.919} & \multicolumn{1}{c}{\textbf{0.938}} \\
\hline
\multirow{2}{*}{$4\times$} & PSNR (dB) & \multicolumn{1}{c}{25.09} & \multicolumn{1}{c}{\underline{25.17}} & \multicolumn{1}{c}{\textbf{25.19}}   \\
                     & SSIM & \multicolumn{1}{c}{\underline{0.798}} & \multicolumn{1}{c}{\textbf{0.803}} & \multicolumn{1}{c}{\underline{0.800}}   \\
\hline \hline
\end{tabular}
\end{minipage}
\end{figure}

We employ two representative inpainting samples used in DIP~\cite{dip01}: \textit{kate} and \textit{vase}. As shown in~\cref{figure8}, our Pool-DIP produces high-quality reconstructions on both samples by restoring fine structures and edges more effectively than the other methods.

\setlength{\textfloatsep}{12pt}




\section{Conclusion}
In this paper, we showed that incorporating multi-scale contextual information via parameter-free pooling operations into non-convolutional architectures can provide a robust and efficient alternative to conventional convolutional DIP models. Our Pool-DIP achieves competitive or improved performance compared to convolution-based architectures. We also provided theoretical analysis supporting the noise-fitting resistance of the proposed architecture. Our findings suggest that non-convolutional architectures equipped with structured pooling mechanisms can serve as effective and lightweight alternatives for DIP. Furthermore, our spectral analysis reveals that Pool-DIP stabilizes the growth of high-frequency components during optimization, which helps mitigate the noise-fitting behavior commonly observed in DIP.
%
%
\bibliographystyle{splncs04}
\bibliography{main}
\end{document}